\documentclass{INTERSPEECH2023}
\interspeechcameraready 

\usepackage{tabularx}

\title{AfriNames: Most ASR models ``butcher" African Names}

\name{
    Tobi Olatunji\textsuperscript{$\dagger$}\thanks{$\dagger$ Equal contribution.}$^{1,2,*}$, Tejumade Afonja$\textsuperscript{$\dagger$}^{3,4,*}$, 
    Bonaventure F. P. Dossou $^{5,6,7,8,*}$, 
    Atnafu Lambebo Tonja$^{9,*}$,
    Chris Chinenye Emezue$^{6,8,10,*}$, 
    Amina Mardiyyah Rufai$^{11,*}$, 
    Sahib Singh$^{12,*}$
}

\address{
    \small
    $^{1}$ Intron Health Inc
    $^{2}$ Georgia Institute of Technology
    $^{3}$ AI Saturdays Lagos
    $^{4}$ CISPA Helmholtz Center for Information Security
    $^{5}$ McGill University
    $^{6}$ Mila Quebec AI Institute
    $^{7}$ Lelapa AI
    $^{8}$ Lanfrica
    $^{9}$ Instituto Politécnico Nacional
    $^{10}$ Technical University of Munich
    $^{11}$ Idiap Research Institute
    $^{12}$ Ford Motor Company
    $^*$Masakhane NLP
}

\email{
    tobi@intron.io, 
    tejumade.afonja@cispa.de, 
    bonaventure.dossou@mila.quebec, 
    atnafu.lambebo@wsu.edu.et, 
    chris.emezue@gmail.com, 
    amina.rufai@idiap.ch, 
    sahibsingh570@gmail.com
}

\begin{document}

\maketitle
 
\begin{abstract}
Useful conversational agents must accurately capture named entities to minimize error for downstream tasks, for example, asking a voice assistant to play a track from a certain artist, initiating navigation to a specific location, or documenting a laboratory result for a patient. However, where named entities such as ``Ukachukwu" (Igbo), ``Lakicia" (Swahili), or ``Ingabire" (Rwandan) are spoken, automatic speech recognition (ASR) models' performance degrades significantly, propagating errors to downstream systems. We model this problem as a distribution shift and demonstrate that such model bias can be mitigated through multilingual pre-training, intelligent data augmentation strategies to increase the representation of African-named entities, and fine-tuning multilingual ASR models on multiple African accents. The resulting fine-tuned models show an 81.5\% relative WER improvement compared with the baseline on samples with African-named entities.
\end{abstract}


\section{Introduction and Motivation}

Automatic Speech Recognition (ASR) powers voice assistants, which use machine learning and other artificial intelligence techniques to automatically interpret and understand spoken languages for conversational purposes. With the advent of breakthroughs such as 
Amazon's Alexa, and Apple's Siri, 
etc., voice assistant technology has increasingly become a widespread technology with diverse applications  \cite{siegert2021speaker}.
However, as these devices gain adoption beyond the demographics of their training data, there is a need for more inclusive and robust AI agents with better spoken language understanding (SLU) and accent recognition capabilities
\cite{desot2019towards, adelani2021masakhaner}\footnote{https://techxplore.com/news/2022-09-effective-automatic-speech-recognition.html}.


Useful conversational agents must accurately capture named entities to minimize errors for downstream tasks. For example, in the command, ``Play Billie Jean by Micheal Jackson", conversational agents need to excel at 3 core tasks: Speech Recognition, Named Entity Recognition, and Entity Linking, to appropriately respond to commands. The ASR component of the system must correctly transcribe the speech, laying a good foundation for Named Entity Recognition (NER) \cite{Nguyen_Yu}, which is, in turn, necessary for effective Entity Linking. 

However, in the command ``Play `Trouble Sleep Yanga Wake Am' by Fela Anikulapo Kuti"\footnote{Fela is one of Africa's most legendary artists} spoken by a Nigerian with a thick Yoruba accent, the phonetic and linguistic variability of the heavily accented speech presents a double dilemma for such systems. Firstly, the heavy accent and tonality can be difficult for the system to recognize, and secondly, the use of out-of-vocabulary words can confuse the model, making it nearly impossible for the system to generate a correct response. Siri responds ``I couldn't find `trouble sleep younger we' by Fela and Kolapo Coochie in your library", effectively ``butchering" \footnote{To "butcher" a name means to mispronounce it, resulting in a significant deviation from the correct pronunciation.} the name, typifying the failures of similar agents on out-of-distribution named entities. More examples in Table \ref{tab:failure_examples}.


We hypothesize that the underrepresentation (and sometimes complete lack of) African named-entities in their training data may partly explain the model bias and eventual ``butchering" of African names by many voice assistants and conversational agents. 


Our contributions are as follows:



    


\begin{enumerate}
    \item We investigate the performance of state-of-the-art (SOTA) ASR models on African named-entities. To do this, we design an effective strategy to evaluate ASR models on speech datasets with no prior NER annotations. Our study highlights the failure of existing SOTA and commercial ASR models on samples with African named-entities
    \item We develop a data augmentation strategy to increase the representation of African-named entities, creating a novel speech corpus rich in African named-entities, and show that by fine-tuning pre-trained models on the augmented accented data, we significantly improve the ability of pre-trained models to recognize African named entities. We open-source the dataset and fine-tuned models\footnote{https://huggingface.co/datasets/tobiolatunji/afrispeech-200}.
    
\end{enumerate}

    


\begin{table*}[t]
\caption{Model behavior examples on native African named entities}
\centering
\small
\begin{tabularx}{\textwidth}{l|X}
\toprule
Model & Sentence\\
\hline
reference & \textbf{Ifeadigo} has been living at \textbf{Kaduna} with his wife \textbf{Chiamaka Orajimetochukwu} \\
\hline

azure & \underline{if you're diego}. \\


aws & \underline{if you did good} has been living at \underline{kaduna} with his wife, \underline{she america or raji mo to} \\



w2v2-lg-960h-lv60-self & \underline{fia digo} has been living at \underline{cadna} with his wife \underline{shi maca orajimo to truco o} \\



w22-lg-xlsr-53-en & \underline{ifia digu} has been living at \underline{kaduna} with his wife \underline{shiamaka orajimo tutruku} \\





whisper-large & \underline{ifeardigun} has been living at \underline{kaduna} with his wife, \underline{shiamaka or rajimu, to chukwu} \\

xlsr-general (Ours) & \underline{ifiadigo} has been living at \underline{kaduna} with his wife \underline{chiamaka orajimotochukwu} \\
Whisper-general (Ours) & \underline{ifeadigo} has been living at \underline{kaduna} with his wife \underline{chiamaka orahjimu tochukwu} \\
\bottomrule
\end{tabularx}
\label{tab:failure_examples}
\end{table*}

\section{Related work}
Developing ASR systems for low-resource languages remains challenging due to the scarcity of training data and resources. As a result, models trained on high-resource languages, such as English, do not perform well on low-resource languages \cite{lepak2021generalisation}. To address this, researchers have proposed several solutions such as cross-lingual representations where the system learns a shared representation for multiple languages \cite{conneau2017word}, data augmentation techniques \cite{feng2021survey}, and fine-tuning ASR model trained on high-resource languages on low-resource languages \cite{anaby2020not}.
Recent SOTA multilingual ASR models such as Whisper \cite{radford2022robust} -- trained on over 680K hours labeled speech samples, including multilingual speech corpora such as Common Voice \cite{commonvoice} -- have significantly improved the ASR landscape, outperforming their monolingual counterparts such as HuBERT \cite{hsu2021hubert}, wavLM \cite{chen2022wavlm}, and wav2vec2 \cite{baevski2020wav2vec} in various downstream tasks. 
Despite these breakthroughs, both open source and commercial ASR systems still exhibit racial bias \cite{koenecke2020racial}, higher error on accented speech \cite{hinsvark2021accented}, and incorrect transcriptions of named entities. Previous studies have highlighted challenges with named entity recognition (NER) for ASR and have investigated various methods to improve NER performance. For instance, French researchers \cite{galliano2009ester} outlined steps for assessing NER in french transcripts of radio broadcasts, while \cite{xiao2021automatic} evaluated Chinese accent ASR on an automatic speech query service (AVQS), highlighting the severe limitations of such systems for Mandarin users with multiple accents. More recently, \cite{mdhaffar2022end, caubriere2020we} attempted to extract semantic information directly from speech signals using a single end-to-end model that learns ASR and NER tasks together. However, none of this work focuses on named entities in African datasets, which presents a new area of research and its unique challenges.

\section{Methodology}
\begin{figure}[h]
\includegraphics[width=\columnwidth]{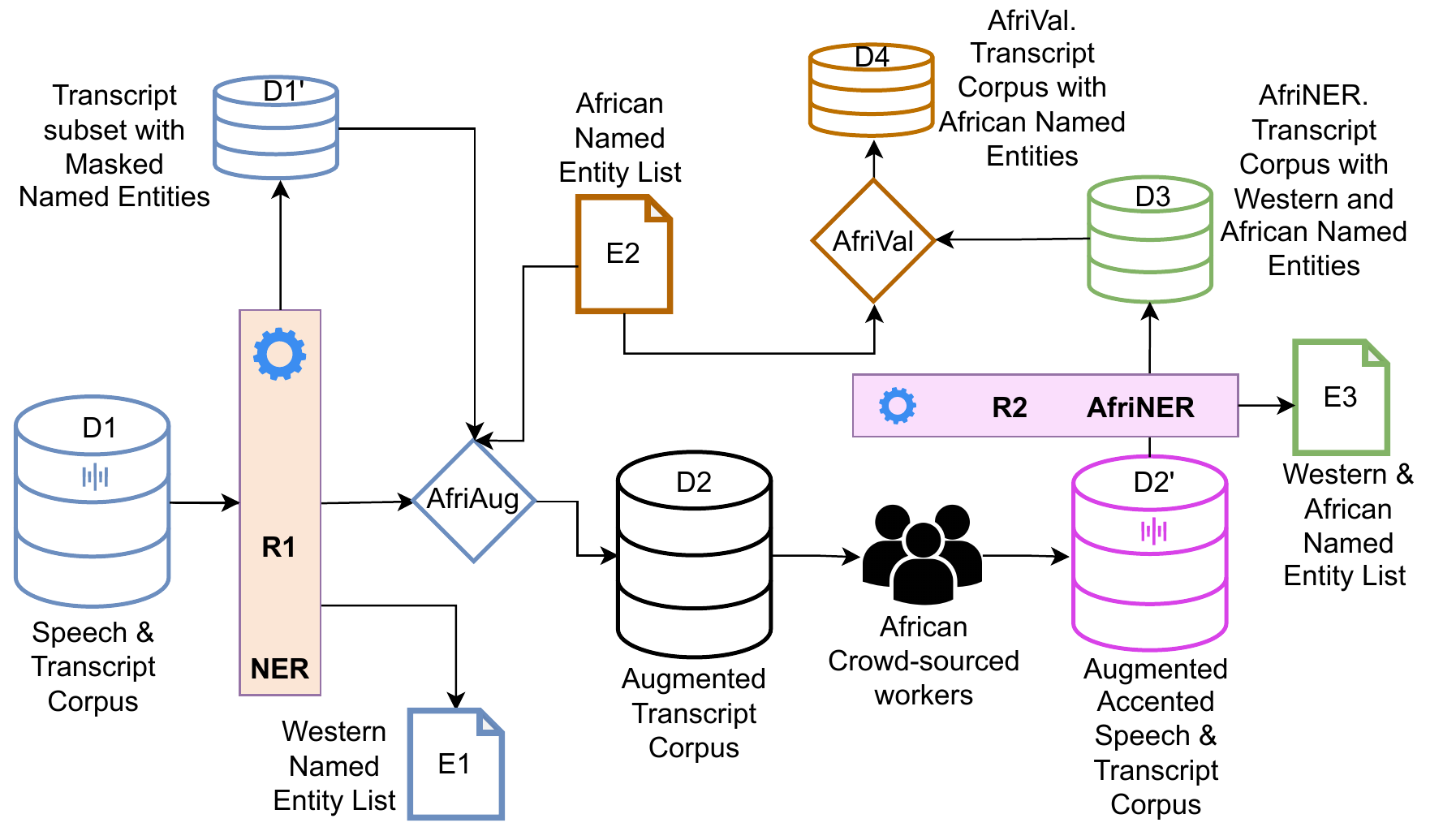} %
\centering
\caption{AfriNames dataset augmentation process.}
\label{fig:method}
\end{figure}

\subsection{African Named-Entity Augmentation Workflow}
\textbf{Western vs African-named entities:} We use the term ``Western named entities" to refer to names that are commonly used in Western cultures and languages, such as Laura and Buenos Aires, and that may not have direct translations in African languages \footnote{Due to the influences of colonization and globalization, many Western names have been adopted in African cultures. Therefore, while these names may not have direct translations in African languages, they can still be used and recognized in African contexts. Our work focus specifically on African named entities that are derived from African languages. }. In contrast, we use the term ``African named entities" to denote names, places, and cultural references that are derived from African languages and cultures, and that may not be commonly used or recognized outside of those contexts.

\textbf{Approach:} 
We address the generalization problem as a domain shift, depicted in Figure \ref{fig:method}. Our initial dataset, denoted as $D_1$, consists of Western audio samples $X^{E_{1}}$ and their corresponding transcripts $Y^{E_{1}}$. We employ a pre-trained named entity recognition (NER) model $R_1$ to extract named entities (NEs) from $Y^{E_{1}}$, resulting in the predominantly Western named entity list $E_1$. To inject African named entities, we mask tokens in randomly selected samples from $Y^{E_{1}}$ that match the entities in $E_1$. This process generates the modified dataset $D_1'$ with modified transcripts $Y'^{E_{1}}$. We then randomly insert tokens from a curated African named-entity list $E_2$ to replace the masked tokens in $Y'^{E_{1}}$, creating an augmented dataset $D_2$ with modified transcripts $Y^{E_{2}}$. These transcripts are sent to African crowd-sourced workers for recording, resulting in a new corpus named $D_2'$ with augmented pairs $\{(X^{E_{2}},Y^{E_{2}})\}$. This novel dataset comprises accented audio samples and augmented transcript pairs, combining distributions from $D_1$ and $D_2$ with Anglo-centric named entities $E_1$ and African named entities $E_2$. Next, we use a specialized NER model $R_2$ to annotate all western and African named entities (called $E_3$) present in $D_2$. Using these NER annotations, we select the subset of $D_2$ with NEs. This NE subset $D_3$ (called AfriNER) contains accented speech $X^{E_{3}}$ and corresponding transcripts $Y^{E_{3}}$ with named entities extracted from both $Y^{E_{1}}$ and $Y^{E_{2}}$. Additionally, using curated African NE list $E_2$, we also filter $Y^{E_{3}}$ to create $D_4$ confirmed to contain African NEs (called AfriVal).

\subsection{Datasets}
In this study, we primarily explore the AfriSpeech-200 dataset, a 200.91 hours novel accented English speech corpus rich with African-named entities, curated for clinical and general domain ASR using the augmentation process described above. 67,577 prompts were recorded by 2,463 unique crowdsourced African speakers from 13 Anglophone countries across sub-Saharan Africa and the United States. The average audio duration was 10.7 seconds (Table \ref{tab: dataset stats}).




\begin{table}[h]
\centering
\small
\caption{AfriSpeech-200 Dataset statistics}
\begin{tabular}{l|l|l|l}
\toprule 
& Train & Dev & Test\\
\midrule
Duration (hrs) & 173.4  &  8.74 & 18.77  \\
\# General domain clips & 21682  &  1407 & 2723  \\
Unique Speakers & 1466   &  247 & 750 \\
Accents & 71 &  45 & 108 \\
\hline
\multicolumn{4}{c}{\textbf{Named Entities Category Counts}}\\
\hline
PER & 11011 &  669 & 1064  \\
ORG & 6322 &  372 & 279 \\
LOC & 3194 &  192 & 526  \\
\bottomrule
\end{tabular}
\label{tab: dataset stats}
\end{table}

\subsection{AfroAug: African Named-Entity Augmentation}\label{section:name-aug}

To increase the representation of African named entities, we start with a corpus $D_1$ using large open-source predominantly western corpora: Wikitext-103 \cite{merity2016pointer} and scrape African entertainment and news websites to increase the representation of African content. We augment this dataset using two main strategies. 
We curate a list $E_2$ of approximately 100k African names using a database of 90,000 African names from \cite{anderson2013using}, 965 Nigerian Igbo names from \cite{okagbue2017personal}, and 1,000 African names obtained from freely available textbooks, online baby name websites, oral interviews, published articles, and online forums like Instagram and Twitter; and African cities list from Wikipedia \footnote{https://en.wikipedia.org/wiki/List\_of\_cities\_in\_Africa\_by\_population}. We augment $D_1'$ in three key steps:
\begin{enumerate}
    \item \textbf{Named-Entity Extraction with NER Models:} We leverage off-the-shelf pre-trained NER models \cite{conneau2019unsupervised} and annotate all named-entities in corpus $Y^{E_{1}}$ to extract the list $E_1$, tokens tagged with [PER], [LOC], or [ORG]. We mask these tokens $e_i \in E_1$ for a randomly sampled subset of transcripts. 
    \item \textbf{Template Selection:} We manually review, select and validate 140 of these sentences where the replacement of masked tokens with African named entities sounds natural and retains meaning in context. These curated sentences with masked tokens are selected as final templates.
    \item \textbf{Named Entity Replacement:} We randomly (uniformly) replace all [LOC] tags with African cities from $E_2$, and all [PER] and [ORG] tags with African names from $E_2$. We repeat this process 200 times to create text corpus $Y^{E_{2}}$ consisting of 28,000 novel augmented transcripts combined with transcripts from $Y^{E_{1}}$ (100,000+ sentences). $Y^{E_{2}}$ is recorded by crowd-sourced workers.
    We sample a subset of users from train/dev/test splits for this work. 
    
\end{enumerate}

A real-world example of $D^{E_{1}}$ is LibriSpeech \cite{panayotov2015librispeech}, a 1,000-hours speech-text dataset from English-only  audiobooks. The resulting ASR model $M_{1}^{E_{1}}$, such as Wav2vec2 \cite{baevski2020wav2vec}, therefore, generalizes poorly to African named entities ${E_{2}}$ (Table \ref{tab:failure_examples}). The pretrained ASR model  $M_{1}^{E_{1}}$ is thus fine-tuned on the new augmented training dataset $D_2'$,  and learns a new mapping $f: X^{E_{2}} \longrightarrow Y^{E_{2}}$ resulting in a more robust model $M_{1}^{E_{2}}$, adapted to the target distribution $D^{E_{2}}$.









\section{Experiments}

\subsection{Benchmarks}\label{section:benchmarks}
We compare SOTA open-source pre-trained ASR models: Whisper \cite{radford2022robust}, Wav2vec2 \cite{baevski2020wav2vec}, XLSR \cite{grosman2021xlsr53-large-english}, Hubert \cite{hsu2021hubert}, and WavLM \cite{chen2022wavlm}, with commercial ASR systems. We refer readers to the respective papers for details on pre-training corpora, model architecture, and hyperparameters. We compare 4 model categories: (1) \textbf{Monolingual Models} pre-trained or fine-tuned exclusively on predominantly western transcripts, western English speech, and western named-entities (2) \textbf{Multilingual Models} pre-trained on transcripts from multiple domains, western and accented speech, but with minimal amounts of African named-entities (3) \textbf{Commercial ASR APIs} (4) \textbf{Ours} finetuned on western and African-named entities paired with audios in accented African English.


\subsection{Fine-tuning}
We select two best-performing open-source models from section \ref{section:benchmarks} and fine-tune them on an accented speech corpus dense with African and western-named entities to achieve robustness to western and African-named entities. We compare pre-trained model performance with fine-tuned checkpoints. Selected model architectures include:
\begin{itemize}
    \item wav2vec2-large-xlsr-53 \cite{grosman2021xlsr53-large-english}: an encoder-decoder architecture with a CNN-based feature extractor, code book, and transformer-based encoder, 378.9M parameters; learning rate of 1e-4.
    \item whisper-medium \cite{radford2022robust}: a decoder-only multi-task architecture, 789.9M parameters; learning rate of 2.5e-4. (We do not fine-tune whisper-large because of computational resource constraints)
\end{itemize}

For each model, we fine-tuned with FP16 \cite{micikevicius2017mixed}, AdamW \cite{loshchilov2019decoupled}, batch size of 16, for 10 epochs, with a linear learning rate decay to zero after a warmup over the first 10\% of iterations. XLSR was trained on a single Tesla T4 GPU with 16GB GPU memory while Whisper was trained on RTX8000 GPU with 48GB GPU memory. Fine-tuning took 24-48 hrs.



\begin{table*}[t]
\tiny
\centering
\caption{WER results on Afrispeech test samples. \textbf{All} is mean WER across all test samples. \textbf{No-NER} is mean WER across samples with NO predicted named entities (NEs). \textbf{AfriNER} is mean WER across all sentences WITH predicted NEs. \textbf{AfriVal} is mean WER across AfriValidated samples. \textbf{char-AfriNER} and \textbf{char-AfriVal} are mean CER on AfriNER and AfriVal respectively. \textbf{char-AfriNER} and \textbf{char-AfriVal} concatenates the NEs in the predicted and reference transcripts.}

\begin{tabular}{l|l|l|l|l|l|l|l|l}
\toprule 
Model & Params & Training or Finetuning data & \multicolumn{4}{c|}{WER}  & \multicolumn{2}{c}{CER} \\
  & & & All (\#2364) & No-NER  (\#1029) & AfriNER  (\#971) & AfriVal  (\#229) &  char-AfriNER & char-AfriVal \\ 
\midrule
\multicolumn{9}{l}{Baseline}\\
\hline
wav2vec2-large-960h & 317M & Monolingual   & 0.641  & 0.565  & 0.696  & 0.802 & 0.861 & 0.986  \\
\midrule
\multicolumn{6}{l}{Monolingual Fine-tuning: Open-Source SOTA pre-trained Models} & \multicolumn{3}{l} {0.718 Monolingual Mean WER}\\
\hline
wav2vec2-large-960h-lv60-self & 317M & Monolingual   & 0.533  & 0.458  & 0.584  & 0.683  & 0.808 & 0.978  \\

hubert-xlarge-ls960-ft & 317M & Monolingual  & 0.562  & 0.487  & 0.613  & 0.701  & 0.803 & 0.986  \\

wavlm-libri-clean-100h-large & 317M & Monolingual   & 0.631  & 0.562  & 0.680  & 0.769  & 0.864 & 0.984  \\


\midrule
\multicolumn{6}{l}{Multilingual Fine-tuning: Open-Source SOTA pre-trained Models } & \multicolumn{3}{l}{0.506 Multilingual Mean WER}\\
\hline
whisper-large & 1550M & Multilingual  & 0.240  & 0.187  & 0.300  & 0.412 & \textbf{0.565} & 0.855 \\
whisper-medium & 769M & Multilingual   & 0.276  & 0.206  & 0.352  & 0.488  & 0.607 & 0.913  \\
wav2vec2-large-xlsr-53-english & 317M & Multilingual   & 0.506  & 0.447  & 0.550  & 0.617  & 0.772 & 0.965  \\

\midrule
\multicolumn{6}{l}{Commercial ASR APIs} & \multicolumn{3}{l}{0.588 Commercial Mean WER}\\
\hline
Azure\cite{azure}  & - & -     & 0.340  & 0.273  & 0.402  & 0.509  & 0.674 & 0.946 \\
GCP\cite{gcp}  & - & -     & 0.534  & 0.464  & 0.603  & 0.700  & 0.827 & 0.991 \\
AWS\cite{aws}  & - & -     & 0.354  & 0.279  & 0.426  & 0.556  & 0.735 & 0.970  \\


\midrule

\multicolumn{6}{l}{AfriSpeech Finetuning (Ours)} & \multicolumn{3}{l}{0.160 AfriSpeech Mean WER} \\
\hline
whisper-medium-AfriSpeech & 769M & Monolingual, AfriSpeech     & \textbf{0.186}  & \textbf{0.172}  & \textbf{0.198}  & \textbf{0.108}  & 0.576 & \textbf{0.704} \\
xlsr-53-english-AfriSpeech & 317M & Monolingual, AfriSpeech     & 0.236  & 0.211  & 0.258  & 0.212  & 0.622 & 0.816  \\

\bottomrule
\end{tabular}

\label{tab:models_benchmarks}
\end{table*}


\subsection{Evaluation}
Word Error Rate (WER) and Character Error Rate (CER) are common metrics for evaluating ASR models. WER measures word errors, CER measures character errors. Lower values are better for both.

\subsubsection{AfriNER: Named-Entity Evaluation}\label{section:ner}
To evaluate ASR performance on named entities (NEs), we need a reliable way to identify samples in $Y_2$ with NEs. Ground truth transcripts $Y_2$ contain $E_1$ and $E_2$ entities, jointly called $E_3$. To extract all samples in $Y_2$ with NEs in $E_3$, we run NER inference on all test samples in $Y_2$ using a specialized performant NER model $R_2$ \footnote{https://huggingface.co/masakhane/afroxlmr-large-ner-masakhaner-1.0\_2.0} from \cite{Adelani2022MasakhaNER2A} that jointly predicts the set of African and western named entities $E_3$. We select test sentences where an entity is detected with confidence (score) greater than 0.8. This seemed to be a reasonable threshold based on ad-hoc analysis. $R_2$ is also able to identify unknown African named entities in $Y_2$ not sourced from $E_2$ (but present in $Y_1$). We denote this subset $Y_3$ (Afri-NER). For each model, we compute WER on corresponding model predictions $Z_3$.




\subsubsection{Sentence-level AfriValidation: African Named-Entity Validation}\label{section:afriVal}
Our primary goal is to evaluate $M_{1}^{E_{1}}$s and $M_{1}^{E_{2}}$ on transcripts with ``African" NEs. To isolate samples with African NEs, we extract the subset of $Y_2$ from the test partition with any NEs from $E_2$ to create the AfriVal subset. Because these sentences are known to contain African NEs, they are Afri-Validated, and guarantee we can reliably evaluate ASR models on predicted transcripts with African NEs. For each model, we compute WER on corresponding model predictions. 


\subsubsection{Character-level AfriValidation}\label{section:char-afriVal}
Since sentence-level WER is impacted by non-NE tokens, we compute CER on NE tokens by isolating them as follows: 1) We run $R_2$ on model predicted transcript $Z_2$ and $Z_3$ to obtain predicted NE tokens with $>0.8$ confidence score. To mitigate the impact of NER errors from $R_2$, for each ground truth and predicted sentence, we concatenate all NER tokens $e_i \in E_3$ from $Y_3$ and all $z_i \in Z_3$ removing all spaces and compute CER.
For selected pre-trained and commercial ASR models $M_{1}^{E_{1}}$, as well as fine-tuned models $M_{1}^{E_{2}}$, we evaluate WER and CER on samples containing one or more named entities and present single run results in Table \ref{tab:models_benchmarks}.



\section{Results and Discussion}
\subsection{African named entities are challenging}
The baseline model in Table \ref{tab:models_benchmarks} demonstrates the dominant trend in our results. WER on all samples (column 4, All) improves by 13.6\% (relative) when samples with named entities are EXCLUDED (column 5, No-NER), worsens by 11.7\% (relative) when samples with named entities (western + African) are isolated (column 6, AfriNER). Performance sinks by 29.7\% (relative) on the subset of Afrivalidated examples (column 7, AfriVal)-- samples with African-named entities from $E2$. This pattern is consistent across all model categories except Ours where we observed a 41.9\% (whisper) and 10.2\% (xlsr) relative WER improvement on AfriVal sentences.

\subsection{Training data bias} 
As shown in Table \ref{tab:models_benchmarks}, multilingual/multitask pre-training outperforms monolingual pre-training/fine-tuning. Multilingual/multitask models \cite{radford2022robust, grosman2021xlsr53-large-english, gulati2020conformer} learn more useful representations, are more linguistically diverse, robust, and generalize better to accented speech when compared with monolingual models fine-tuned on datasets (e.g. Librispeech  \cite{panayotov2015librispeech} and Switchboard \cite{godfrey1992switchboard}) with predominantly western NEs and western accents. After fine-tuning on AfriSpeech with African NEs and accented speech, our best model, whisper-medium improves on the baseline by 81.5\% compared to 16.4\% for the pre-trained model.


\subsection{Multilingual pretraining is insufficient}  
Despite extensive pretraining on 680k hours of multilingual data (90 languages), the fine-tuned model outperforms the pre-trained model by 77.9\% (relative). Our results demonstrate that multilingual/multi-task pretraining is inadequate as these SOTA models make several mistakes with African-named entities. Fine-tuning results show that our approach is effective in mitigating bias in these large models. 



\subsection{Character-Level analysis} 
When named entities are isolated as described in Section \ref{section:afriVal}, we observe that our fine-tuned whisper-medium model worsens by 1.9\% (relative) in comparison to the pre-trained whisper-large model (column 8, char-AfriNER). This may be due to the significantly higher number of parameters in whisper-large generalizing better to certain named entities.
However, when evaluated on the Afrivalidated dataset (column 9, char-AfriVal), our fine-tuned whisper-medium model outperforms both pre-trained whisper-large and medium models (relative gain of 17.7\%, and 22.9\% respectively). These results further support our claim that the presence of African-named entities is crucial for achieving better performance in ASR models. 

\subsection{Use of language models} 
Table \ref{tab:failure_examples} shows some of the difficulties with commercial APIs where a language model (LM) is likely used to rescore the raw ASR transcript. This is especially destructive for African-named entities. Because these named entities (e.g.``Ifeadigo") are missing from LM training data, where the probability of sequences with African NEs is effectively zero, and such transcripts are downranked by the algorithm in favor of more likely tokens like ``Diego" as seen in the example in Table \ref{tab:failure_examples}. Prediction score thresholds may also be in use under the hood in these commercial systems, limiting the ASR output where confidence is low resulting in truncated output as seen in Table \ref{tab:failure_examples}. 

\section{Conclusion}
Automatic speech recognition (ASR) for African-named entities is a challenging task for most state-of-the-art (SOTA) ASR models including those trained with multilingual data and multitask objectives. We demonstrate that this bias can be mitigated by fine-tuning these models on accented speech corpora rich in African-named entities, shifting the distribution for robustness in the African context.








\bibliographystyle{IEEEtran}
\bibliography{mybib}

\end{document}